\documentclass[conference]{IEEEtran}
\IEEEoverridecommandlockouts
\usepackage{cite}
\usepackage{multirow}
\usepackage{array}
\usepackage{amsmath,amssymb,amsfonts}
\usepackage{algorithmic}
\usepackage{graphicx}
\usepackage{textcomp}
\usepackage{makecell}
\usepackage{xcolor}
\usepackage{hyperref}
\def\BibTeX{{\rm B\kern-.05em{\sc i\kern-.025em b}\kern-.08em
    T\kern-.1667em\lower.7ex\hbox{E}\kern-.125emX}}

\makeatletter
\newcommand{\twocolumnfootnotefullwidth}[1]{%
  \begingroup
  \renewcommand{\thefootnote}{}
  \footnotetext{%
    \noindent\hspace*{-1em}\rule{0.3\linewidth}{0.4pt}\\[0.2em] 
    \vspace*{-1em}
    \noindent\footnotesize #1
  }
  \endgroup
}
\makeatother

\begin{document}

\title{FOSCU: Feasibility of Synthetic MRI Generation via Duo-Diffusion Models for Enhancement of 3D U-Nets in Hepatic Segmentation}

\author{
\IEEEauthorblockN{Youngung Han\textsuperscript{*\dag}}
\IEEEauthorblockA{\textit{Seoul National University} \\
yuhan@snu.ac.kr}
\and
\IEEEauthorblockN{Kyeonghun Kim\textsuperscript{\dag}}
\IEEEauthorblockA{\textit{OUTTA} \\
kyeonghun.kim@outta.ai}
\and
\IEEEauthorblockN{Seoyoung Ju\textsuperscript{\dag}}
\IEEEauthorblockA{\textit{Sangmyung University} \\
202115055@sangmyung.kr}
\and
\IEEEauthorblockN{Yeonju Jean\textsuperscript{\dag}}
\IEEEauthorblockA{\textit{Ewha Womans University} \\
ahxlzjt@ewhain.net}
\and
\IEEEauthorblockN{Minkyung Cha\textsuperscript{*}}
\IEEEauthorblockA{\textit{Seoul National University} \\
cmk4911@snu.ac.kr}
\and
\IEEEauthorblockN{Seohyoung Park\textsuperscript{\dag}}
\IEEEauthorblockA{\textit{Ewha Womans University} \\
03nobel@ewhain.net}
\and
\IEEEauthorblockN{Hyeonseok Jung}
\IEEEauthorblockA{\textit{Chung-Ang University} \\
mjmk0820@cau.ac.kr}
\and
\IEEEauthorblockN{Nam-Joon Kim\textsuperscript{*\ddag}}
\IEEEauthorblockA{\textit{Seoul National University} \\
knj01@snu.ac.kr}
\and
\IEEEauthorblockN{Woo Kyoung Jeong}
\IEEEauthorblockA{\textit{Samsung Medical Center, Sungkyunkwan University School of Medicine} \\
jeongwk@skku.edu}
\and
\IEEEauthorblockN{Ken Ying-Kai Liao}
\IEEEauthorblockA{\textit{NVIDIA} \\
kenyingkail@nvidia.com}
\and
\IEEEauthorblockN{Hyuk-Jae Lee\textsuperscript{*}}
\IEEEauthorblockA{\textit{Seoul National University} \\
hjlee@capp.snu.ac.kr}
}

\maketitle

\twocolumnfootnotefullwidth{* Seoul National University \quad {\dag} OUTTA \quad {\ddag} Corresponding author}


\begin{abstract}
Medical image segmentation faces fundamental challenges including restricted access, costly annotation, and data shortage to clinical datasets through Picture Archiving and Communication Systems (PACS). These systemic barriers significantly impede the development of robust segmentation algorithms. To address these challenges, we propose FOSCU, which integrates Duo-Diffusion, a 3D latent diffusion model with ControlNet that simultaneously generates high-resolution, anatomically realistic synthetic MRI volumes and corresponding segmentation labels, and an enhanced 3D U-Net training pipeline. Duo-Diffusion employs segmentation-conditioned diffusion to ensure spatial consistency and precise anatomical detail in the generated data. Experimental evaluation on 720 abdominal MRI scans shows that models trained with combined real and synthetic data yield a mean Dice score gain of 0.67\% over those using only real data, and achieve a 36.4\% reduction in Fréchet Inception Distance (FID), reflecting enhanced image fidelity.
\end{abstract}




\section{Introduction}
Medical image segmentation serves as a cornerstone of modern healthcare, enabling precise diagnosis, and disease monitoring across diverse clinical applications \cite{1_cai2020review, 2_ma2024segment}. Accurate identification of anatomical features and pathological areas within medical imaging is essential, as segmentation accuracy significantly influences clinical decision-making and patient outcomes. However, the development of robust segmentation models faces significant challenges, particularly in specialized domains such as liver imaging, where anatomical complexity and data scarcity create substantial barriers to achieving optimal performance \cite{3_wang2018fully,4_bougourzi2025recent}.

Liver segmentation poses significant difficulties in medical imaging analysis, primarily because of various complexities related to anatomy and imaging techniques. The liver exhibits highly variable morphology across patients, with shape variations that complicate automated segmentation approaches \cite{5_manjunath2024automated,6_ghobadi2025challenges,7_ansari2022practical}. 

Medical image segmentation development continues to face persistent challenges that hinder clinical progress. These include: (1) \textbf{restricted data access}: clinical data stored in PACS \cite{8_eichelberg2020cybersecurity} are subject to privacy regulations and institutional constraints, creating bottlenecks for algorithm development; (2) \textbf{data shortage}: the limited prevalence of certain pathologies and anatomical variations reduce the diversity of available training datasets, especially in hepatic imaging where lesion heterogeneity complicates data collection; and (3) \textbf{costly annotation}: annotating medical images requires specialized radiological expertise to define anatomical boundaries and pathological features accurately, making the process significantly more labor-intensive and costly than general computer vision tasks. 


\begin{figure}[t!]
\centering
\includegraphics[width=\columnwidth]{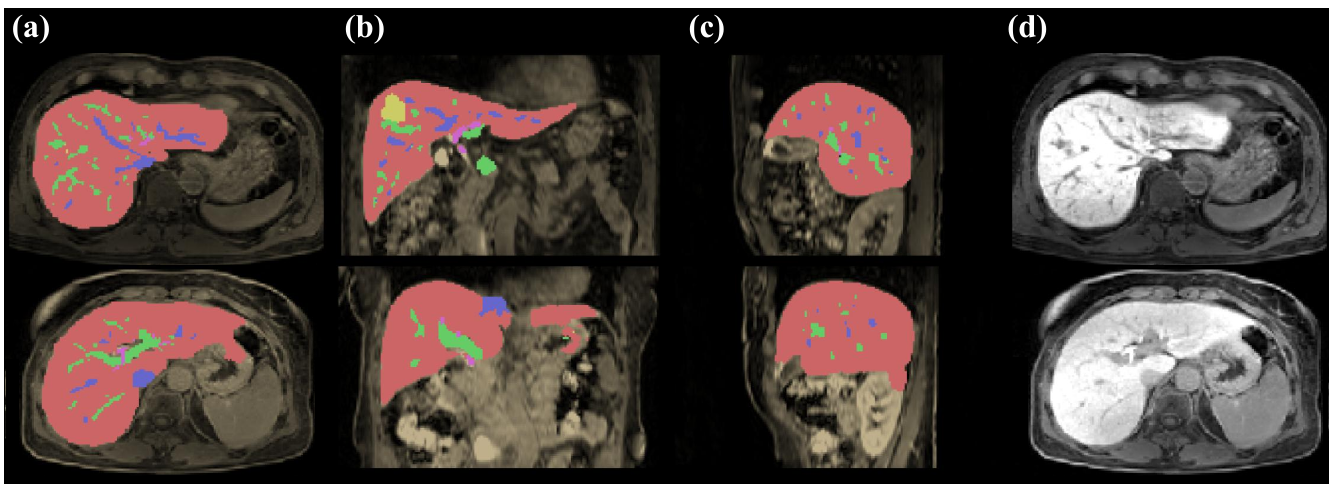}
\caption{Examples of generated high-resolution synthetic abdominal MRI volumes with corresponding segmentation conditions overlaid. (a) Axial, (b) Coronal, and (c) Sagittal views showing segmentation masks inferred by a 3D U-Net trained on data generated with Duo-Diffusion. (d) Axial slices extracted from the generated 3D synthetic MRI volume.}
\label{fig1}
\end{figure}

\begin{figure*}[t]
\centering
\includegraphics[width=\textwidth]{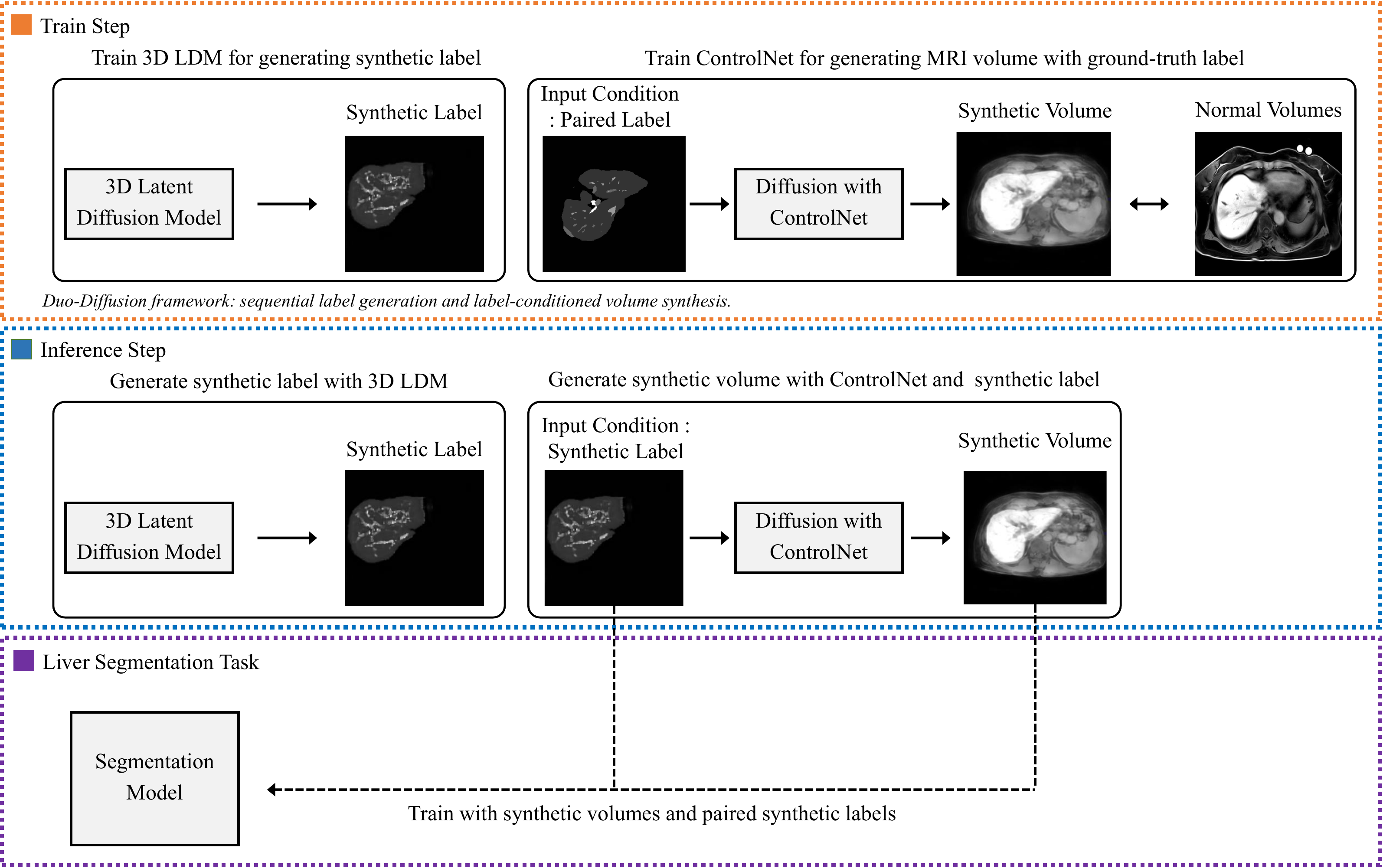}
\caption{Overview of the proposed FOSCU framework. The workflow consists of three steps: (1) training the Duo-Diffusion architecture, which involves first training a 3D LDM to generate synthetic labels and then training a ControlNet to synthesize MRI volumes conditioned on paired labels, (2) performing inference with the pretrained Duo-Diffusion models to sequentially generate synthetic labels and corresponding synthetic volumes, and (3) leveraging the generated paired synthetic data to train a liver segmentation model capable of segmenting liver structures from synthetic MR images.
}
\label{fig2}
\end{figure*}

To address these limitations, we introduce Duo-Diffusion, a novel 3D latent diffusion architecture within the FOSCU framework specifically tailored for medical image augmentation. Motivated by prior studies exploring a range of generative models \cite{9_ddpm,10_ldm,11_gan,12_stylegan,13_mirza2014conditional,14_muller2023multimodal,15_zhang2023adding}, our approach generates high-resolution synthetic MRI volumes along with their corresponding segmentation labels in a unified pipeline. This dual-generation strategy first synthesizes anatomically consistent segmentation masks and then uses them as spatial conditioning inputs to create realistic abdominal MRI volumes. By providing diverse, high-fidelity synthetic data aligned with expert-defined labels, Duo-Diffusion effectively enhances the training of segmentation models, improving anatomical accuracy and overall performance as demonstrated in Fig.~\ref{fig1}.

\begin{figure*}[t]
\centering
\includegraphics[width=\textwidth]{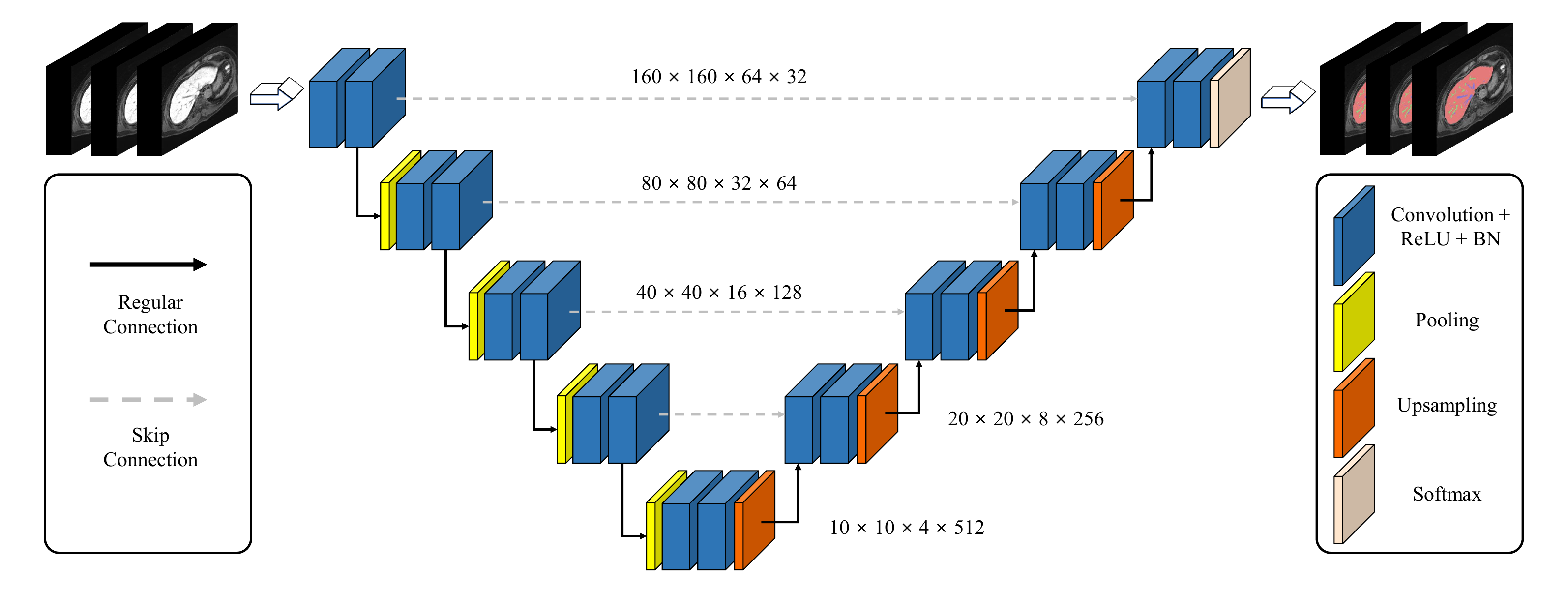}
\caption{The segmentation model in the proposed FOSCU framework is based on a standard 3D U-Net architecture. The network takes 3D synthetic abdominal MRI volumes as input and outputs liver segmentation masks. The numbers in each block indicate the feature map dimensions \(H\times W\times D\times C\), corresponding to height, width, depth, and channels.}
\label{fig3}
\end{figure*}

\section{Methods}

\subsection{Data Preparation}
Anonymized abdominal MRI scans from 720 patients at Samsung Medical Center were employed for this study. These scans, acquired over several years on multiple MRI systems, were filtered to retain only those captured during the hepatobiliary phase, which provides optimal liver contrast. Each patient's two-dimensional DICOM slices were visually reviewed to confirm image quality, and manual annotations were generated for four structures—liver, portal vein, hepatic vein, and tumor—serving as the ground truth for segmentation. All individual planar slices were then consolidated to generate a single volumetric NIfTI file per patient.

To accelerate model training and safeguard patient privacy, we extracted a cubic region of interest (ROI) with dimensions of (160, 160, 64) voxels from each NIfTI volume. Using ITK-SNAP, we manually inspected all 720 volumes to identify the lesion's centroid in three‐dimensional space and centered the ROI on these coordinates. The cropped ROIs were normalized via min–max range scaling to the $[0,1]$ interval. 

\subsection{Duo-Diffusion}
We developed Duo-Diffusion, a two-stage diffusion framework designed to generate high-quality synthetic medical volumes from label conditions. Duo-Diffusion combines a 3D LDM\cite{10_ldm} for liver segmentation labels and a 3D LDM with ControlNet\cite{15_zhang2023adding} for MRI volumes, enabling both label synthesis and volume generation within a unified pipeline.
In the first stage, the first 3D LDM learns to generate synthetic segmentation labels from noise by compressing input volumes of size (8, 24, 24, 12) into a compact latent space and reconstructing them through a decoder. The training objective for the diffusion process in this stage is defined as:
\begin{equation}
\mathcal{L}_{\text{diffusion}} = \mathbb{E}_{\mathcal{E}_l(l),\, \epsilon_l \sim \mathcal{N}(0,1),\, t} \left[ \left\| \epsilon_l - \epsilon_{l, \theta}(z_l, t, t) \right\|_2^2 \right]
\end{equation}
where $\epsilon_l$ denotes the noise added to the latent label representation, and $\epsilon_{l, \theta}$ is the noise predicted by the model parameterized by $\theta$.
In the second stage, the second 3D LDM learns to generate synthetic volumes using the same architectural approach, with ControlNet providing conditional guidance from segmentation labels. ControlNet is trained to generate MRI volumes conditioned on real segmentation labels. The training loss for this conditional generation is given by:
\begin{equation}
\mathcal{L}_{c} = \mathbb{E}_{\mathcal{E}(x),\, \{\epsilon, \epsilon_l\} \sim \mathcal{N}(0,1),\, t} \left[ \left\| \epsilon - \epsilon_{\theta}(z_t, t, c_f(z_l)) \right\|_2^2 \right]
\end{equation}
where $\epsilon$ is the noise added to the latent image representation, $\epsilon_{\theta}$ is the model's prediction conditioned on both the noisy image and the label latent, and $c_f$ is the task-specific condition.
This Duo-Diffusion approach ensures precise anatomical consistency between segmentation masks and image volumes while maintaining high fidelity. As illustrated in Fig.~\ref{fig2} (orange box), the framework integrates both diffusion processes sequentially to produce paired synthetic datasets for training downstream segmentation models.

\subsection{3D U-Nets}
For liver segmentation in our FOSCU framework, we employ several 3D U-Net architectures that extend the encoder–decoder paradigm to volumetric medical imaging data \cite{16_ronneberger2015u}. Figure~\ref{fig3} illustrates the base 3D U-Net in our segmentation pipeline.

We utilize multiple 3D U-Net variants, including ResUNet\cite{17_resunet} (with residual shortcuts), WideResUNet\cite{18_wideresunet} (with expanded channel dimensions), DynUNet\cite{19_nnunet} (with an adaptive architecture), and VNet\cite{20_vnet} (with PReLU activations). All models were trained using Dice loss\cite{21_sudre2017generalised} to maximize voxel-wise overlap:

\begin{equation}
\mathcal{L}_{\text{Dice}} = 1 - \frac{2\sum_{i} p_i g_i + \varepsilon}{\sum_{i} p_i + \sum_{i} g_i + \varepsilon}
\end{equation}

where $p_i$ and $g_i$ denote the prediction and ground truth, respectively, and $\varepsilon$ is a small fixed constant (e.g., $10^{-6}$) added to ensure numerical stability by preventing division by zero.





\section{Experiments}
All experiments were conducted on a single NVIDIA A100 80 GB GPU using MONAI 1.3.2, MONAI Generative 0.2.3, and Python 3.10 \cite{22_monai}. For dataset partitioning, we allocated 504 patients for training, 72 for validation, and 144 for testing. We used the AdamW optimizer from PyTorch for all the gradient descent processes.

\subsection{Duo-Diffusion Training}
The Duo-Diffusion framework comprises two separate 3D LDMs. Each model consists of a VAE\cite{23_autoencoder} and a diffusion component\cite{9_ddpm}, with the second LDM enhanced by ControlNet\cite{15_zhang2023adding}. Both 3D LDMs share identical hyperparameters and training procedures but were trained for distinct generation tasks.

First, the VAE uses reconstruction loss with a KL-divergence weight of \(10^{-7}\) and a learning rate of \(10^{-6}\). It was trained for 2,000 epochs, and this phase required approximately three days on the A100 GPU.

Next, the diffusion component uses \(L_2\) loss~\cite{9_ddpm} with a learning rate of \(10^{-5}\). It was trained for 4,000 epochs, and this phase took approximately two days on the A100 GPU.


Finally, ControlNet is integrated into the Duo-Diffusion framework and uses the same dataset. It uses a learning rate of \(10^{-5}\) and targets enhancing the sharpness and clarity of specified regions within the generated synthetic volumes. It was trained for an additional 5,000 epochs, and this final stage required approximately two days on the A100 GPU.


\subsection{3D U-Nets Training}
%

Multiple variants of 3D U-Net are evaluated for binary liver segmentation and multi-class segmentation, as detailed in Table~\ref{tab2}. The binary task uses binary cross-entropy loss, and the multi-class task uses categorical cross-entropy loss. Each model is trained with an early stopping patience of 10 and a learning rate of \(10^{-4}\). All models were trained on an NVIDIA A100 GPU, and training for each model was completed in approximately one day.



\section{Results}

\subsection{High-Resolution Synthetic MRI}
\begin{figure}[htbp]
\centering
\includegraphics[width=1\columnwidth]{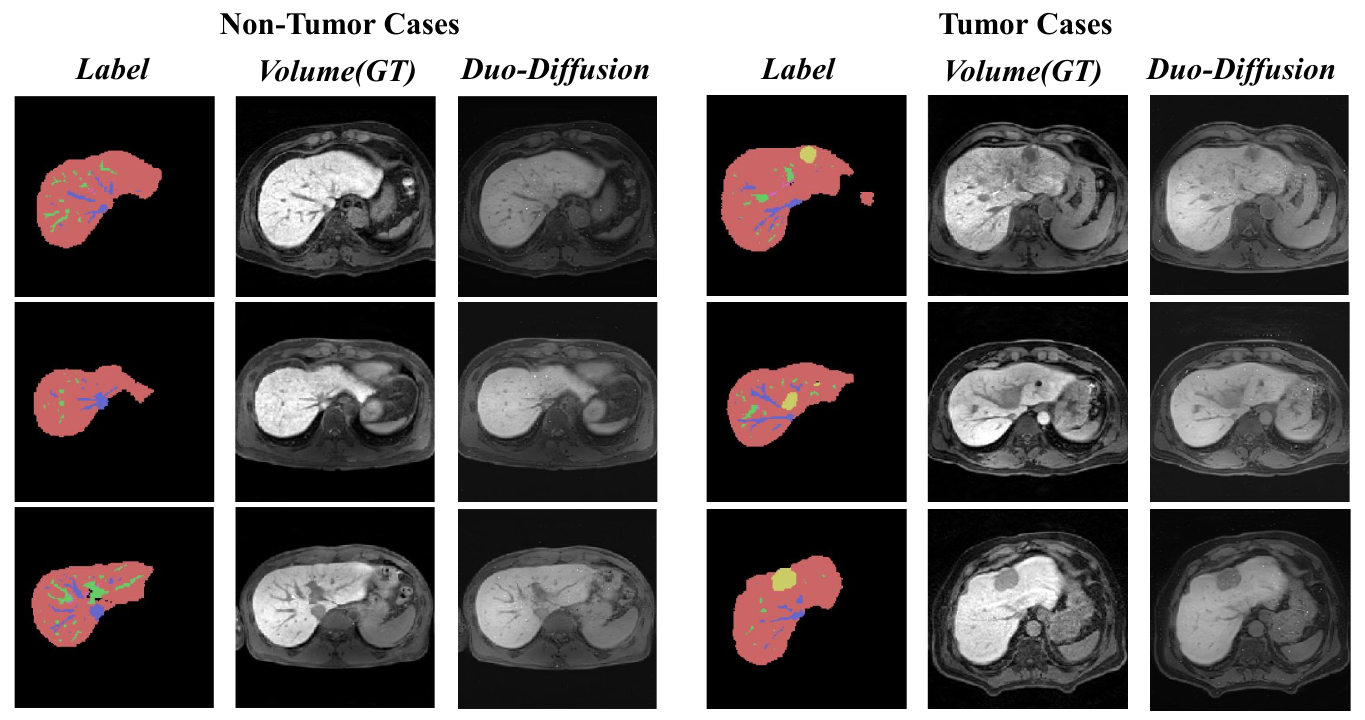}
\caption{Representative examples of synthetic abdominal MRI generation. The left columns show non-tumor cases and the right columns show tumor (HCC) cases. Each example includes the ground truth label, the original volume, and the synthetic volume generated by Duo-Diffusion.}
\label{fig4}
\end{figure}

We evaluated the generative models using the Fréchet Inception Distance (FID)\cite{24_fid}. As shown in Table~\ref{tab1}, our Duo-Diffusion method achieved superior performance with an FID score of 28.31 compared to other 3D-LDM variants. Representative examples of the synthetic MRI quality for tumor and non-tumor cases were shown in Fig.~\ref{fig4}. Additionally, a standard VAE baseline achieved an FID of 21.67. These results show that our method outperforms competing 3D-LDM variants, indicating enhanced visual quality and representation accuracy.

\renewcommand{\arraystretch}{1.4}
\begin{table}[htbp]
\caption{Generative Model Performance Comparison}
\label{tab1}
\centering
\begin{tabular}{
    |>{\centering\arraybackslash}m{2.5cm}
    |>{\centering\arraybackslash}m{0.9cm}
    |>{\centering\arraybackslash}m{0.9cm}
    |>{\centering\arraybackslash}m{0.9cm}
    |>{\centering\arraybackslash}m{0.9cm}|}
\hline
\textbf{Method} & \textbf{FID ↓ (Ax.)} & \textbf{FID ↓ (Sag.)} & \textbf{FID ↓ (Cor.)} & \textbf{FID ↓ (Avg.)} \\
\hline
\makecell{3D-LDM {\scriptsize(w/ VQVAE)}} & 67.17 & 63.43 & 62.96 & 64.52 \\
\hline
\makecell{3D-LDM {\scriptsize(w/ VAE)}} & 40.69 & 37.34 & 37.83 & 38.62 \\
\hline
\makecell{\textbf{Duo-Diffusion}} & \textbf{29.27} & \textbf{27.62} & \textbf{28.04} & \textbf{28.31} \\
\hline
\end{tabular}
\end{table}


\subsection{Liver Segmentation Performance}

We evaluated segmentation performance by training several CNN architectures on datasets augmented with our synthetic data. As shown in Table~\ref{tab2}, 3D U-Nets trained on combined real and synthetic datasets consistently outperformed those trained on real data alone, achieving improved Dice coefficients. For liver-only segmentation, in which all non-background regions are relabeled as a single liver class, we observed consistent performance gains across models. In the more challenging multi-class segmentation task—targeting liver, portal vein, hepatic vein, and tumor—the inclusion of synthetic data yielded even more pronounced improvements, with accuracy increases of up to 0.85\% across architectures. 

\renewcommand{\arraystretch}{1.3}
\begin{table}[htbp]
\caption{Liver Segmentation Performance Results}
\label{tab2}
\centering
\begin{tabular}{
    |>{\centering\arraybackslash}m{1.6cm}
    |>{\centering\arraybackslash}m{1.55cm}
    |>{\centering\arraybackslash}m{0.72cm}
    |>{\centering\arraybackslash}m{1.35cm}
    |>{\centering\arraybackslash}m{1.55cm}|
}
\hline
\textbf{CNN Model} & \textbf{Segmentation} & \textbf{Real} & \textbf{\makecell{Real\\+ Synthesis}} & \textbf{Improvement} \\
\hline
\multirow{2}{*}{U-Net} & Liver-Only & 0.9650 & 0.9662 & \textbf{+0.12}\% \\
\cline{2-5}
& Multi-Class & 0.6968 & 0.7014 & \textbf{+0.66}\% \\
\hline
\multirow{2}{*}{ResUNet} & Liver-Only & 0.9633 & 0.9634 & \textbf{+0.01}\% \\
\cline{2-5}
& Multi-Class & 0.6601 & 0.6652 & \textbf{+0.77}\% \\
\hline
\multirow{2}{*}{WideResUNet} & Liver-Only & 0.9657 & 0.9649 & -0.08\% \\
\cline{2-5}
& Multi-Class & 0.6961 & 0.7020 & \textbf{+0.85}\% \\
\hline
\multirow{2}{*}{DynUNet} & Liver-Only & 0.9692 & 0.9701 & \textbf{+0.09}\% \\
\cline{2-5}
& Multi-Class & 0.7301 & 0.7340 & \textbf{+0.53}\% \\
\hline
\multirow{2}{*}{VNet} & Liver-Only & 0.9614 & 0.9618 & \textbf{+0.04}\% \\
\cline{2-5}
& Multi-Class & 0.6643 & 0.6679 & \textbf{+0.54}\% \\
\hline
\multirow{2}{*}{\makecell{Overall Mean\\DICE}} & Liver-Only & 0.9649 & 0.9653 & \textbf{+0.04}\% \\
\cline{2-5}
& Multi-Class & 0.6895 & 0.6941 & \textbf{+0.67}\% \\
\hline
\end{tabular}
\end{table}

\section{Discussion and Conclusion}


This study introduces the FOSCU framework, an approach for addressing data shortage challenges in medical imaging via our Duo-Diffusion architecture. Duo-Diffusion is a dual-generation diffusion model that simultaneously synthesizes high-resolution MRI volumes and the corresponding segmentation labels for liver segmentation. Through this synthetic data generation approach, we achieved substantial improvements in 3D U-Net-based liver segmentation performance.

However, some liver segmentation architectures showed only modest performance gains, likely due to the dataset’s limited size. Comprehensive validation across larger, multi-institutional datasets remains essential \cite{25_lits,26_chaos}. In future study, we will conduct a comparative evaluation of our Duo-Diffusion methodology against alternative generative models and augmentation techniques \cite{27_chen2022generative,28_kazerouni2023diffusion}. 


\section*{Acknowledgment}


This work was supported by the Next Generation Semiconductor Convergence and Open Sharing System, and by the Institute of Information \& Communications Technology Planning \& Evaluation (IITP) under the Artificial Intelligence Semiconductor Support Program to Nurture the Best Talents (IITP-2023-RS-2023-00256081), funded by the Korea government (MSIT).

\newpage

\bibliographystyle{IEEEtran}
\bibliography{references}

\end{document}